\documentclass[11pt,a4paper]{article}
\usepackage[hyperref]{acl2019}
\usepackage{times}
\usepackage{latexsym}

\usepackage{url}

\usepackage{graphicx}
\usepackage{amsfonts}
\usepackage{dsfont}
\usepackage{amsmath}
\usepackage{xcolor}
\usepackage[inline]{enumitem}
\usepackage{xspace}

\aclfinalcopy 
\def\aclpaperid{1437} 

\newcommand\todo[1]{}
\newcommand\todoy[1]{}
\newcommand\cut[1]{}
\newcommand\revision[1]{#1}

\def\NAME{LCD}

\title{A Cross-Domain Transferable Neural Coherence Model}

\author{Peng Xu$^1$, 
  Hamidreza Saghir$^1$,
  Jin Sung Kang$^1$, 
  Teng Long$^1$,  
  Avishek Joey Bose\thanks{\ Work done while the author was an intern at Borealis AI.} $^{,1,2}$, \\
  {\bf Yanshuai Cao$^1$,
  Jackie Chi Kit Cheung$^{1,2,3}$}\\
  $^1$Borealis AI \\
  \texttt{\small \{peng.z.xu,hamidreza.saghir,jinsung.kang,leo.long,yanshuai.cao\}@borealisai.com}\\ 
  $^2$McGill University\\
  \texttt{\small \{joey.bose,jcheung\}@cs.mcgill.ca} \\
  $^3$Canada CIFAR Chair, Mila}

\date{}

\begin{document}
\maketitle
\begin{abstract}
Coherence is an important aspect of text quality and is crucial for ensuring its readability. One important limitation of existing coherence models is that training on one domain does not easily generalize to unseen categories of text. Previous work \cite{li2017neural} advocates for generative models for cross-domain generalization, because for discriminative models, the space of incoherent sentence orderings to discriminate against during training is prohibitively large. In this work, we propose a local discriminative neural model with a much smaller negative sampling space that can efficiently learn against incorrect orderings. The proposed coherence model is simple in structure, yet it significantly outperforms previous state-of-art methods on a standard benchmark dataset on the Wall Street Journal corpus, as well as in multiple new challenging settings of transfer to unseen categories of discourse on Wikipedia articles.\footnote{Code and datasets are available at: \url{github.com/BorealisAI/cross_domain_coherence}.}

\end{abstract}



\section{Introduction}

Coherence is a discourse property that is concerned with the logical and semantic organization of a passage, such that the overall meaning of the passage is expressed fluidly and clearly. It is an important quality measure for text generated by humans or machines, and modelling coherence can benefit many applications, including summarization, question answering \cite{verberne2007evaluating}, essay scoring \cite{miltsakaki2004evaluation, burstein2010using} and text generation \cite{park2015expressing, kiddon2016globally, holtzman2018learning}.

The ability to generalize to new domains of text is desirable for NLP models in general. Besides the practical reason of avoiding costly retraining on every new domain, for coherence modelling, we would also like our model to make decisions based on the semantic relationships between sentences, rather than simply overfit to the structural cues of a specific domain.

The standard task used to test a coherence model in NLP is sentence ordering, for example, to distinguish between a coherently ordered list of sentences and a random permutation thereof.
Earlier work focused on feature engineering, drawing on theories such as Centering Theory \cite{grosz1995centering} and Rhetorical Structure Theory \cite{thompson1987rhetorical} to propose features based on local entity and lexical transitions, as well as more global concerns regarding topic transitions \cite{elsner2007unified}. With the popularization of deep learning, the focus has shifted towards specifying model architectures, including a number of recent models that rely on distributed word representations used in a deep neural network \cite{li2017neural, nguyen2017neural, logeswaran2018sentence}.

One key decision which forms the foundation of a model is whether it is discriminative or generative. Discriminative models depend on contrastive learning; they use automatic corruption methods to generate incoherent passages of text, then learn to distinguish coherent passages from incoherent ones. By contrast, generative approaches aim at maximising the likelihood of the training text, which is assumed to be coherent, without seeing incoherent text or explicitly incorporating coherence into the optimization objective. 

It has been argued that neural discriminative models of coherence are prone to overfitting on the particular dataset and domain that they are designed for \cite{li2017neural}, possibly due to the expressive nature of functions learnable by a neural network. Another potential problem for discriminative models raised by Li and Jurafksy is that there are $n!$ possible sentence orderings for a passage with $n$ sentences, thus the sampled negative instances can only cover a tiny proportion of this space, limiting the performance of such models. There is thus an apparent association between discriminative models and high performance on a specific narrow domain.

We argue in this paper that there is, in fact, nothing inherent about discriminative models that cause previous systems to be brittle to domain changes. We demonstrate a solution to the above problems by combining aspects of previous generative and discriminative models to produce a system that works well in both in-domain and cross-domain settings, despite being a discriminative model overall.

Our method relies on two key ideas. The first is to reexamine the operating assumption of previous work that a global, passage-level model is necessary for good performance. While it is true that coherence is a property of a passage as a whole, capturing long-term dependencies in sequences remains a fundamental challenge when training neural networks in practice \cite{trinh2018learning}. On the other hand, it is {\em plausible} that much of global coherence can be decomposed into a series of local decisions, as demonstrated by foundational theories such as Centering Theory. Our hypothesis in this work is that there remains much to be learned about local coherence cues which previous work has not fully captured and that these cues make up an essential part of global coherence. We demonstrate that such is the case.

Our model thus take neighbouring pairs of sentences as inputs, for which the space of negatives is much smaller and can therefore be effectively covered by sampling other sentences in the same document. Surprisingly, adequately modelling local coherence alone significantly outperforms previous approaches, and furthermore, local coherence captures text properties that are domain-agnostic, generalizing much better in open-domain settings to unseen categories of text. \todoy{Maybe say a few words about why local information is more domain-agnostic}

Our second insight is that the superiority of previous generative approaches in cross-domain settings can be effectively incorporated into a discriminative model as a pre-training step. We show that generatively pre-trained sentence encoders enhance the performance of our discriminative local coherence model.

We demonstrate the effectiveness of our approach on the Wall Street Journal (WSJ) benchmark dataset, as well as on three challenging new evaluation protocols using different categories of articles drawn from Wikipedia that contain increasing levels of domain diversity. We show that our discriminative model significantly outperforms strong baselines on all datasets tested. Finally, hypothesis testing shows that the coherence scores from our model have a significant statistical association with the ``rewrite'' flag for regular length Wikipedia articles, demonstrating that our model prediction aligns with human judgement of text quality.

To summarize, our contributions are:
\vspace{-.2cm}
\begin{enumerate}
\itemsep0em
\item We correct the misconception that discriminative models cannot generalize well for cross-domain coherence scoring, with a novel local discriminative neural model.
\item We propose a set of cross-domain coherence datasets with increasingly difficult evaluation protocols.
\item Our new method outperforms previous methods by a significant margin on both the previous closed domain WSJ dataset as well as on all open-domain ones, setting the new state-of-the-art for coherence modelling.
 \item Even with the simplest sentence encoder, averaged GloVe, our method frequently outperforms previous methods, while it can gain further accuracy by using stronger encoders.
\end{enumerate}

\section{Related Work}

\newcite{barzilay2008modeling} introduced the entity grid representation of a document, which uses the local syntactic transitions of entity mentions to model discourse coherence. Three tasks for evaluation were introduced for evaluation: discrimination, summary coherence rating, and readability assessment.
Many models were proposed to extend this model \cite{eisner2011extending, feng2012extending, guinaudeau2013graph}, including
 models relying on HMMs \cite{louis2012coherence} to model document structure.

Driven by the success of deep neural networks, many neural models were proposed in the past few years.
\newcite{li2014model} proposed a neural clique-based discriminative model to compute the coherence score of a document by estimating a coherence probability for each clique of $L$ sentences.
\newcite{nguyen2017neural} proposed a neural entity grid model with convolutional neural network that operates over the entity grid representation.
\newcite{mohiuddin2018coherence} extended this model for written asynchronous conversations.
Both methods rely on hand-crafted features derived from NLP preprocessing tools to enhance the original entity grid representation. We take a different approach to feature engineering in our work, focusing on the effect of supervised or unsupervised pre-training.

\newcite{li2017neural} was the first work to use generative models to model coherence and proposed to evaluate the performance of coherence models in an open-domain setting.
Most recently, \newcite{logeswaran2018sentence} used an RNN based encoder-decoder architecture to model the coherence which can also be treated as the generative model.
One obvious disadvantage of generative models is that they maximize the likelihood of training text but never see the incoherent text. In other words, to produce a binary classification decision about coherence, such a generative model only sees data from one class. As we will demonstrate later in the experiments, this puts generative models at a disadvantage comparing to our local discriminative model.

\section{Background: Generative Coherence Models}

To understand the advantages of our local discriminative model, we first introduce the previous global passage-level generative coherence models. We will use ``passage'' and ``document'' interchangeably in this work, since all the models under consideration work in the same way for a full document or a passage in document.

Generative models are based on the idea that in a coherent passage, subsequent sentences should be predictable given their preceding sentences, and vice versa.
Let us denote the corpus as $\mathcal{C}=\{d_k\}_{k=1}^N$, which consists of $N$ documents, with each document $d_k$ comprised of a sequence of sentences $\{s_i\}$. Formally, generative coherence models are trained using a log-likelihood objective as follows (with some variations according to the specific model):
\begin{equation}
\max_{\theta} \sum_{d \in \mathcal{C}}\sum_{s \in d} \log p(s \vert c_s;\theta),
\end{equation}
\noindent where $c_s$ is the context of the sentence $s$ and $\theta$ represents the model parameters. $c_s$ can be chosen as the next or previous sentence \cite{li2017neural}, or all previous sentences \cite{logeswaran2018sentence}.

There are two hidden assumptions behind this maximum likelihood approach to coherence. First, it assumes that conditional log likelihood is a good proxy for coherence. Second, it assumes that training can well capture the long-range dependencies implied by the generative model.

Conditional log likelihood essentially measures the compressibility of a sentence given the context; i.e., how predictable $s$ is given $c_s$. However, although incoherent next sentence is generally not predictable given the context, the inverse is not necessarily true. In other words, a coherent sentence does not need to have high conditional log-likelihood, as log likelihood can also be influenced by other factors such as fluency, grammaticality, sentence length, and the frequency of words in a sentence. Second, capturing long-range dependencies in neural sequence models is still an active area of research with many challenges \cite{trinh2018learning}, hence there is no guarantee that maximum likelihood learning can faithfully capture the inductive bias behind the first assumption.

\section{Our Local Discriminative Model}
\revision{We propose the local coherence discriminator model (LCD) whose} operating assumption is that the global coherence of a document can be well approximated by the average of coherence scores between consecutive pairs of sentences. Our experimental results later will validate the appropriateness of this assumption. For now, this simplification allows us to cast the learning problem as discriminating consecutive sentence pairs $(s_i,s_{i+1})$ in the training documents (assumed to be coherent) from incoherent ones $(s_i, s')$ (negative pairs to be constructed). 

\paragraph{Training objective:}
Formally, our discriminative model $f_\theta(.,.)$ takes a sentence pair and returns a score. The higher the score, the more coherent the input pair. Then our training objective is:
\begin{equation}
\mathcal{L}(\theta)\!= \!\sum_{d \in \mathcal{C}}\sum_{s_i \in d}\mathop{\mathbb{E}}_{p(s'|s_i)}\!\left[ L(f_\theta(s_i, s_{i+1}), f_\theta(s_i, s'))\right]
\end{equation}
where $\mathop{\mathbb{E}}_{p(s'|s_i)}$ denotes expectation with respect to negative sampling distribution $p$ which could be conditioned on $s_i$; and $L(.,.)$ is a loss function that takes two scores, one for a positive pair and one for a negative sentence pair.

\paragraph{Loss function:}
The role of the loss function is to encourage $f^+ = f_\theta(s_i, s_{i+1})$ to be high while $f^- = f_\theta(s_i, s')$ to be low. Common losses such as margin or log loss can all be used. Through experimental validation, we found that margin loss to be superior for this problem. Specifically, $L$ takes on the form:
$L(f^+, f^-) = \max(0, \eta - f^+ + f^-)$
where $\eta$ is the margin hyperparameter.

\paragraph{Negative samples:}
Technically, we are free to choose any sentence $s'$ to form a negative pair with $s_i$. However, because of potential differences in genre, topic and writing style, such negatives might cause the discriminative model to learn cues unrelated to coherence. Therefore, we only select sentences from the same document to construct negative pairs. Specifically, suppose $s_i$ comes from document $d_k$ with length $n_k$, then $p(s'|s_i)$ is a uniform distribution over the $n_k\!-\!1$ sentences $\{s_j\}_{j~\neq~i}$ from $d_k$. For a document with $n$ sentences, there are $n\!\!-\!\!1$ positive pairs, and $(n\!-\!1)\!*\!(n\!-\!2)/2$ negative pairs. It turns out that the quadratic number of negatives provides a rich enough learning signal, while at the same time, is not too prohibitively large to be effectively covered by a sampling procedure.
In practice, we sample a new set of negatives each time we see a document, hence after many epochs, we can effectively cover the space for even very long documents.
Section~\ref{sec_details} discusses further details on sampling.


\subsection{Model Architecture}
\label{model}

\begin{figure}[ht]
\label{bigram}
\begin{center}
 \includegraphics[height=2.2in]{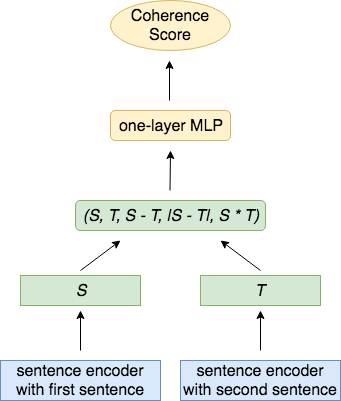}
\end{center}
\caption{Generic architecture for our proposed model.} 
\label{bigram_model}
\end{figure}

The specific neural architecture that we use for $f_{\theta}$ is illustrated in Figure~\ref{bigram_model}. We assume the use of some pre-trained sentence encoder, which is discussed in the next section.

Given an input sentence pair, the sentence encoder maps the sentences to real-valued vectors $S$ and $T$. We then compute the concatenation of the following features:
\begin{enumerate*}[label=(\arabic*)]
\item concatenation of the two vectors $(S,T)$;
\item element-wise difference $S-T$;
\item element-wise product $S*T$;
\item absolute value of element-wise difference $\lvert S-T\rvert$.
\end{enumerate*}
The concatenated feature representation is then fed to a one-layer MLP to output the coherence score.

In practice, we make our overall coherence model bidirectional, by training a forward model with input $(S, T)$ and a backward model with input $(T, S)$ with the same architecture but separate parameters.
The coherence score is then the average from the two models.

\subsection{Pre-trained Generative Model as the Sentence Encoder}
\label{sec:encoder}
Our model can work with any pre-trained sentence encoder, ranging from the most simplistic average GloVe \cite{pennington2014glove} embeddings to more sophisticated supervised or unsupervised pre-trained sentence encoders \cite{conneau2017supervised}.
As mentioned in the introduction, since generative models can often be turned into sentence encoder, generative coherence model can be leveraged by our model to benefit from the advantages of both generative and discriminative training, similar to \cite{kiros2015skip,peters2018deep}. After initialization, we freeze the generative model parameters to avoid overfitting. 

In Section~\ref{sec_exp}, we will experimentally show that while we do benefit from strong pre-trained encoders, the fact that our local discriminative model improves over previous methods is independent of the choice of sentence encoder.

\section{Experiments}
\label{sec_exp}
\subsection{Evaluation Tasks}
Following \newcite{nguyen2017neural} and other previous work, we evaluate our models on the discrimination and insertion tasks. Additionally, we evaluate on the paragraph reconstruction task in open-domain settings, in a similar manner to \newcite{li2017neural}.

In the {\em discrimination} task, a document is compared to a random permutation of its sentences, and the model is considered correct if it scores the original document higher than
the permuted one. Twenty permutations are used in the test set in accordance with previous work.

In the {\em insertion} task, we evaluate models based on their ability to find the correct position of a sentence that has been removed from a document.
To measure this, each sentence in a given document is relocated to every possible position.
An insertion position is selected for which the model gives the highest coherence score to the document. The insertion score is then computed as the average fraction of sentences per document reinserted into their original position.

In the {\em reconstruction} task, the goal is to recover the original correct order of a shuffled paragraph given the starting sentence. We use beam search to drive the reconstruction process, with the different coherence models serving as the selection mechanism for beam search. We evaluate the performance of different models based on the rank correlation achieved by the top-1 reconstruction after search, averaged across different paragraphs.

For longer documents, since a random permutation is likely to be different than the original one at many places, the discrimination task is easy. Insertion is much more difficult since the candidate documents differ only by the position of one sentence. Reconstruction is also hard because small errors accumulate.

\subsection{Datasets and Protocols}
\paragraph{Closed-domain:}
The single-domain evaluation protocol is done on the Wall Street Journal (WSJ) portion of Penn Treebank (Table~\ref{wsj}), similar to previous work \cite{nguyen2017neural}\footnote{Since the preprocessing pipeline is different, we generate the random permutations by ourselves. \todoy{Say a few words about difference in preprocessing?}}.

\paragraph{Open-domain:}
\newcite{li2017neural} first proposed open-domain evaluation for coherence modelling using Wikipedia articles, but did not release the dataset\footnote{A version of preprocessed data consisting of integer indices is available online, but it is not usable without important preprocessing details.\todoy{A version of preprocessed data consisting of integer indices is available online, but it is not usable without important preprocessing details.}}.

Hence, we create a new dataset based on Wikipedia and design three cross-domain evaluation protocols with increasing levels of difficulty.
Based on the ontology defined by DBpedia\footnote{http://mappings.dbpedia.org/server/ontology/classes/}, we choose seven different categories under the domain {\bf\em Person} and three other categories from irrelevant domains.
We parse all the articles in these categories and extract paragraphs with more than $10$ sentences to be used as the passages for training and evaluation.
The statistics of this dataset is summarized in Table~\ref{wiki_datasets}. 
The three settings with increasing level of hardness are as follows:
\begin{enumerate}
\itemsep0em
\item {\bf Wiki-A(rticle)} randomly split all paragraphs of the seven categories under {\bf\em Person} into training part and testing part;
\item {\bf Wiki-C(ategory)} hold out paragraphs in one category from {\bf\em Person} for evaluation and train on the remaining categories in {\bf\em Person};
\item {\bf Wiki-D(omain)} train on all seven categories in {\bf\em Person}, and evaluate on completely different domains, such as Plant, Institution, CelestialBody, and even WSJ.
\end{enumerate}

Wiki-A setting is essentially the same protocol as the open domain evaluation as the one used in \cite{li2017neural}. Importantly, there is no distribution drift (up to sampling noise) between training and testing. Thus, this protocol only tests whether the coherence model is able to capture a rich enough set of signal for coherence, and does not check whether the learned cues are specific to the domain, or generic semantic signals. For example, cues based on style or regularities in discourse structure may not generalize to different domains. Therefore, we designed the much harder Wiki-C and Wiki-D to check whether the coherence models capture cross-domain transferrable features. In particular, in the Wiki-D setting, we even test whether the models trained on {\bf\em Person} articles from Wikipedia generalize to WSJ articles.
\begin{table}[ht]
 \small
\centering
\begin{tabular}{l| c | c | c}
\hline
\textbf{Domain} & \textbf{Category} & \textbf{\# Paras} & \textbf{Avg. \# Sen.} \\ \hline
Person & Artist & 9553 & 11.87 \\
& Athlete & 23670 & 12.26 \\
& Politician & 2420 & 11.62 \\
& Writer & 3310 & 11.83 \\
& MilitaryPerson & 6428 & 11.90 \\
& OfficeHolder & 6578 & 11.54 \\
& Scientist & 2766 & 11.77 \\ \hline
Species & Plant & 3100 & 12.26 \\
Organization & Institution & 5855 & 11.58 \\
Place & CelestialBody & 414 & 11.55 \\ \hline
\end{tabular}
\caption{Statistics of the Wiki Dataset.}
\label{wiki_datasets}
\end{table}

\subsection{Baselines}
We compared our proposed model \revision{\NAME} against two document-level discriminative models:
(1) Clique-based discriminator {\bf Clique-Discr.} \cite{li2014model,li2017neural} with window size $3$ and $7$.
(2) Neural entity grid model {\bf Grid-CNN} and {\bf Extended Grid-CNN} \cite{nguyen2017neural};
And three generative models:
(3) {\bf Seq2Seq} \cite{li2017neural};
(4) {\bf Vae-Seq2Seq} \cite{li2017neural}\footnote{As the authors did not release their code, so we tried our best at replicating their model. Some important implementation details are missing from \cite{li2017neural}, so we cannot guarantee exactly the same setup as in \cite{li2017neural}.}
;
(5) {\bf LM}, an RNN language model, and use the difference between conditional log likelihood of a sentence given its preceding context, and the marginal log likelihood of the sentence.
All the results are based on our own implementations except {\bf Grid-CNN} and {\bf Extended Grid-CNN}, for which we used code from the authors. 

We compare these baselines to our proposed model \revision{with three different encoders}:
\begin{enumerate}
 \itemsep0em
 \item \revision{{\bf \NAME-G}}: use averaged GloVe vectors \cite{pennington2014glove} as the sentence representation;
 \item \revision{{\bf \NAME-I}}: use pre-trained InferSent \cite{conneau2017supervised} as the sentence encoder;
 \item \revision{{\bf \NAME-L}}: apply max-pooling on the hidden state of the language model to get the sentence representation.
\end{enumerate}

\subsection{Results on Domain-specific Data}
\vspace{-.25cm}
\begin{table}[ht]
\centering
\begin{tabular}{l | c | c}
\hline
& \bf{Discr.} & \bf{Ins.} \\ \hline
Clique-Discr. (3) & 70.91 & 11.53\\
Clique-Discr. (7) & 70.30 & 5.01\\
Grid-CNN & 85.57 (85.13) & 23.12 \\
Extended Grid-CNN & 88.69 (87.51) & 25.95 \\ \hline
Seq2Seq & 86.95 & 27.28 \\
Vae-Seq2Seq & 87.01& 26.73\\
LM & 86.50 & 26.33 \\ \hline
\revision{\NAME-G} & 92.51 & 30.30 \\
\revision{\NAME-I} & 94.54 & 32.34 \\
\revision{\NAME-L} & {\bf 95.49} & {\bf 33.79} \\ \hline
\end{tabular}
\caption{Accuracy of {\bf Discr}imination and {\bf Ins}ertion tasks evaluated on WSJ. For Grid-CNN and Extended Grid-CNN, the numbers outside brackets are taken from the corresponding paper, and numbers shown in the bracket are based on our experiments with the code released by the authors.}
\label{wsj}
\end{table}

We first evaluate the proposed models on the Wall Street Journal (WSJ) portion of Penn Treebank (Table~\ref{wsj}).
Our proposed models perform significantly better than all other baselines, even if we use the most na\"ive sentence encoder, {\em i.e.}, averaged GloVe vectors.
Among all the sentence encoders, {\bf LM} trained on the local data in an unsupervised fashion performs the best, better than {\bf InferSent} trained on a much larger corpus with supervised learning.
In addition, combining the generative model {\bf LM} with our proposed architecture as the sentence encoder improves the performance significantly over the generative model alone.

\subsection{Results on Open-Domain Data}
\begin{table}[h]
\centering
\begin{tabular}{l | c}
\hline
Clique-Discr. (3) & 76.17 \\
Clique-Discr. (7) & 73.86 \\ \hline
Seq2Seq & 86.63 \\
Vae-Seq2Seq & 82.40\\
LM & 93.83 \\ \hline
\revision{\NAME-G} & 91.32 \\
\revision{\NAME-I} & 94.01 \\
\revision{\NAME-L} & {\bf 96.01} \\ \hline
\end{tabular}
\caption{Accuracy of discrimination task under Wiki-A}
\label{wiki_a}
\end{table}

\begin{table*}[ht]
\small
\centering
\begin{tabular}{l | c c c c c c c | c}
\hline
Model & Artist & Athlete & Politician & Writer & MilitaryPerson & OfficeHolder & Scientist & Average \\ \hline
Clique-Discr. (3) & 73.01 & 68.90 & 73.82 & 73.28 & 72.86& 73.74 & 74.56 & 72.88 \\
Clique-Discr. (7) & 71.26 & 66.56 & 73.72& 72.01& 72.67& 72.62 & 71.86 & 71.53 \\ \hline
Seq2Seq & 82.72 & 73.45 & 84.88 & 85.99 & 81.40 & 83.25 & 85.27 & 82.42 \\
Vae-Seq2Seq &  82.58 & 74.14 &  84.70 & 84.94 &  81.07 &  82.66 &  85.09 & 82.17 \\
LM & 88.18 & 78.79 & 88.95 & 90.68 & 87.02 & 87.35 & 91.92 & 87.56 \\ \hline
\revision{\NAME-G} & 89.66 & 86.06 & 90.98 & 90.26 & 89.23 & 89.86 & 90.64 & 89.53 \\
\revision{\NAME-I} & 92.14 & 89.03 & 93.23 & 92.07 & 91.63 & 92.39 & 93.03 & 91.93 \\
\revision{\NAME-L} & {\bf 93.54} & {\bf  90.13} & {\bf 94.04} & {\bf 93.68} & {\bf 93.20} & {\bf 93.01} & {\bf 94.81} & {\bf 93.20}\\ \hline
\end{tabular}
\caption{Accuracy of discrimination task under Wiki-C setting.}
\label{wiki_c}
\end{table*}

\begin{table*}[ht]
\centering
\small
\begin{tabular}{l | c c c c | c}
\hline
Model & Plant & Institution & CelestialBody & WSJ & Average \\ \hline
Clique-Discr. (3) & 66.14 & 66.51 & 60.38 & 64.71 & 64.44 \\
Clique-Discr. (7) & 65.47 & 69.14 & 61.44 & 66.66 & 65.68\\ \hline
Seq2Seq & 82.58 & 80.86 & 69.44 & 74.62 & 76.88 \\
Vae-Seq2Seq &  81.90 &  78.00 &  69.10 & 73.27 & 75.57  \\
LM & 81.88 & 83.82 & 74.78 & 79.78 & 80.07 \\ \hline
\revision{\NAME-G} & 86.57 & 86.10 & 79.16 & 82.51 & 83.59 \\
\revision{\NAME-I} & {\bf 89.07} & 88.58 & 80.41 & {\bf 83.27} & 85.33 \\
\revision{\NAME-L} & 88.83 & {\bf 89.46} & {\bf 81.31} & 82.23 & {\bf 85.48} \\ \hline
\end{tabular}
\caption{Accuracy of discrimination task under Wiki-D setting.}
\label{wiki_d}
\end{table*}

We next present results in the more challenging open-domain settings. Tables~\ref{wiki_a},~\ref{wiki_c},~and \ref{wiki_d} present results on the discriminative task under the Wiki-A, Wiki-C, Wiki-D settings. We do not report results of the neural entity grid models, since these models heavily depend on rich linguistics features from a preprocessing pipeline, but we cannot obtain these features on the Wiki datasets with high enough accuracy using standard preprocessing tools. \todoy{I don't understand what this sentence says. We didn't do because the code doesn't work right? JC: I tried to clarify. Please check and correct.}
As in the closed-domain setting, our proposed models outperform all the baselines for almost all tasks even with the averaged GloVe vectors as the sentence encoder.
Generally, \revision{{\bf \NAME-L}} performs better than \revision{{\bf \NAME-I}}, but their performances are comparable under Wiki-D setting. This result may be caused by the fact that {\bf InferSent} is pre-trained on a much larger dataset in a supervised way, and generalizes better to unseen domains.

As the Wiki-A setting is similar to the open-domain setting proposed by \newcite{li2017neural}, and we also have similar observations as stated in their papers. The generative models perform quite well under this setting and applying them on top of our proposed architecture as the sentence encoder further enhances their performances, as illustrated in Table~\ref{wiki_a}.
However, as observed in Tables~\ref{wiki_c} and~\ref{wiki_d}, the generative models do not generalize as well into unseen categories, and perform even worse in unseen domains.
We emphasize that a protocol like Wiki-A or similar setup considered in \newcite{li2017neural} is insufficient for evaluating open domain performance. Because difficulties in open domain coherence modelling lie not only in the variety of style and content in the dataset, but also in the fact that training set cannot cover all potential variation there is in the wild, making cross domain generalization a critical requirement. \todo{I don't understand the previous sentence. Please rewrite it?} \todoy{how about now?}
\todoy{describe how there's little drop from Wiki-A to Wiki-C, and to Wiki-D}

\subsection{Paragraph Order Reconstruction Results}
\vspace{-.2cm}
\begin{table}[h]
    \begin{tabular}{l|cc}
    \hline
    Model & Wiki-D ({\small CelestialBody}) & Wiki-A \\ \hline
    Seq2Seq & 0.2104 & 0.2119 \\
    LM & 0.1656 & 0.1420 \\
    \revision{\NAME-I} & \textbf{0.2507} & 0.2744 \\
    \revision{\NAME-L} & 0.2326 & \textbf{0.2900} \\
    \hline
    \end{tabular}
    \caption{Kendall's tau for re-ordering on Wiki-A/-D}
    \label{sentence_reordering}
\end{table}
As shown by the discrimination and insertion tasks, Seq2Seq and LM are the stronger baselines, so for paragraph reconstruction, we compare our method to them, on two cross domain settings, the simpler Wiki-A and the harder Wiki-D. We report the reconstruction quality via Kendall's tau rank correlation in Table~\ref{sentence_reordering}, which shows that our method is superior by a significant margin.

\subsection{Hyperparameter Setting and Implementation Details}
\label{sec_details}
In this work, we search through different hyperparameter settings by tuning on the development data of the WSJ dataset, then apply the same setting across all the datasets and protocols. The fact that one set of hyperparameters tuned on the closed-domain setting works across all protocols, including open-domain ones, demonstrates the robustness of our method.


The following hyperparameter settings are chosen: Adam optimizer \cite{kingma2014adam} with default settings and learning rate $0.001$, and no weight decay; the number of hidden state $d_h$ for the one-layer MLP as $500$, input dropout probability $p_i$ as $0.6$, hidden dropout probability $p_h$ as $0.3$; the margin loss was found to be superior to log loss, and margin of $5.0$ was selected. In addition, we use early-stopping based on validation accuracy in all runs.

Furthermore, during training, every time we encounter a document, we sample $50$ triplets $(s_i, s_{i+1}, s')$'s, where $(s_i, s_{i+1})$'s form positive pairs while $(s_i, s')$'s form negative pairs.
So effectively, we resample sentences so that documents are trained for the same number of steps regardless of the length.
For all the documents including the permuted ones, we add two special tokens to indicate the start and the end of the document.
\todoy{For wikipedia, our training is done on paragraph levels, so sampling is always from the same paragraph, this is important, and different from WSJ}
\subsection{Analysis}
\subsubsection{Ablation Study}
\label{sec_ablation}
To better understand how different design choices affect the performance of our model, we present the results of an ablation study using variants of our best-performing models in Table~\ref{ablation}. The protocol used for this study is Wiki-D with CelestialBody and Wiki-WSJ, the two most challenging datasets in all of our evaluations.

The first variant uses a unidirectional model instead of the default bidirectional mode with two separately trained models. The second variant only uses the concatenation of the two sentence representations as the features instead of the full feature representation described in Section~\ref{model}.

\begin{table}[h]
\centering
\begin{tabular}{l | c | c }
\hline
Model  & CelestialBody & Wiki-WSJ \\ \hline
\revision{\NAME-L} & 81.31 & 82.23  \\
no bidirectional & 80.33 & 82.30 \\
no extra features & 79.28 & 79.84 \\ \hline
\end{tabular}
\caption{Ablation study: {\bf Discr.} under Wiki-D}
\label{ablation}
\end{table}

As we can see, even our ablated models still outperform the baselines, though performance drops slightly compared to the full model. This demonstrates the effectiveness of our general framework for modelling coherence.

\subsubsection{Effect of Sample Coverage}
\vspace{-.4cm}
\begin{figure}[ht]
\label{bigram}
\begin{center}
 \includegraphics[width=.85\columnwidth]{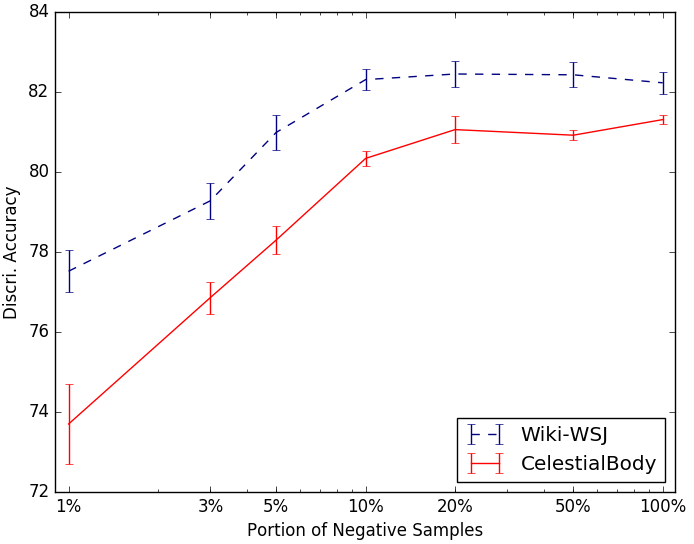}
\end{center}
\vspace{-.4cm}
\caption{Discrimination accuracy on CelestialBody and Wiki-WSJ with different portions of all valid samples. The \emph{x} axis is in log-scale.}
\label{sample}
\end{figure}
Previous work raised concerns that negative sampling cannot effectively cover the space of negatives for discriminative learning \cite{li2017neural}.
Fig.~\ref{sample} shows that for our local discriminative model, there is a diminishing return when considering greater coverage beyond certain point ($20\%$ on these datasets). Hence, our sampling strategy is more than sufficient to provide good coverage for training.


%

\subsubsection{Comparison with Human Judgement}
To evaluate how well our coherence model aligns with human judgements of text quality, we compare our coherence score to Wikipedia's article-level ``rewrite'' flags. This flag is used for articles that do not adhere to Wikipedia's style guidelines, which could be due to other reasons besides text coherence, so this is an imperfect proxy metric. Nevertheless, we aim to demonstrate a potential correlation here, because carelessly written articles are likely to be both incoherent and in violation of style guidelines. This setup is much more challenging than previous evaluations of coherence models, as it requires the comparison of two articles that could be on very different topics.

For evaluation, we want to verify whether there is a difference in average coherence between articles marked for rewrite and articles that are not. We select articles marked with an article-level rewrite flag from Wikipedia, and we sample the non-rewrite articles randomly. We then choose articles that have a minimum of two paragraphs with at least two sentences. We use our model trained for the Wiki-D protocol, and average its output scores per paragraph, then average these paragraph scores to obtain article-level scores. This two-step process ensures that all paragraphs contribute roughly equally to the final coherence score. We then perform a one-tailed $t$-test for the mean coherence scores between the rewrite and no-rewrite groups. 

We find that among articles of a typical length between 2,000 to 6,000 characters (Wikipedia average length c. 2,800 characters), the average coherence scores are $0.56$ (marked for rewrite) vs. $0.79$ (not marked) with a p-value of $.008$. For longer articles of 8,000 to 14,000 characters, the score gap is smaller ($0.60$ vs $0.64$), and p-value is $0.250$. It is possible that in the longer marked article, only a subportion of the article is incoherent, or that other stylistic factors play a larger role, which our simple averaging does not capture well.

\section{Conclusion}
\vspace{-.1cm}
In this paper, we examined the limitations of two general frameworks for coherence modelling; i.e.~, passage-level discriminative models and generative models.
We propose a simple yet effective local discriminative neural model which retains the advantages of generative models while addressing the limitations of both kinds of models.
Experimental results on a wide range of tasks and datasets demonstrate that the proposed model outperforms previous state-of-the-art methods significantly and consistently on both domain-specific and open-domain datasets.



\section*{Acknowledgements}
We thank all the anonymous reviewers for their valuable inputs.

\bibliography{ref}
\bibliographystyle{acl_natbib}

\end{document}